\documentclass[10pt,twocolumn,letterpaper]{article}

\usepackage{iccv}
\usepackage{times}
\usepackage{epsfig}
\usepackage{graphicx}
\usepackage{amsmath}
\usepackage{amssymb}

\usepackage{ifthen}
\usepackage{color}
\usepackage{bigstrut}
\usepackage{multirow}
\usepackage{mathtools}
\usepackage[normalem]{ulem}
\usepackage{algorithm}
\usepackage{algpseudocode}
\usepackage{subcaption}

\newcommand{\myfootnote}[2]{{%
  \let\thempfn\relax% Remove footnote number printing mechanism
  \footnotetext[0]{$^#1$\emph{#2}}% Print footnote text
}}

\DeclarePairedDelimiter{\norm}{\lVert}{\rVert}
\usepackage[pagebackref=true,breaklinks=true,colorlinks,citecolor=blue,linkcolor=blue,bookmarks=false]{hyperref}
\usepackage{booktabs}

\definecolor{frenchblue}{rgb}{0.0, 0.45, 0.73}
\definecolor{gray}{rgb}{0.5,0.5,0.5} 
\definecolor{green}{rgb}{0, 0.4, 0} 
\definecolor{orange}{rgb}{1, 0.5, 0} 	
\definecolor{mahogany}{rgb}{0.75, 0.25, 0.0}
\definecolor{purple}{rgb}{0.6, 0, 0.6}
\definecolor{darkgreen}{rgb}{0, 0.4, 0.4} 
\definecolor{black}{rgb}{0, 0, 0}
\definecolor{teal}{rgb}{0.0, 0.5, 0.5}
\definecolor{aaaa}{rgb}{0.55, 0.1, 0.7}
\definecolor{red}{rgb}{1.0, 0, 0}

\newboolean{revising}
\setboolean{revising}{false}
\ifthenelse{\boolean{revising}}
{
	\newcommand{\fuen}[1]{\textcolor{blue}{[FuEn]: #1}}

} {
	\newcommand{\fuen}[1]{#1}

}

\iccvfinalcopy % *** Uncomment this line for the final submission

\def\iccvPaperID{4635} % *** Enter the ICCV Paper ID here

\ificcvfinal\fi

\begin{document}

%%%%%%%%% TITLE
\title{VMCML: Video and Music Matching via Cross-Modality Lifting}

\iffalse
\author{Yi-Shan Lee\\
National Tsing Hua University\\
{\tt\small lys13.ee06@nycu.edu.tw}
\and
Wei-Cheng Tseng\\
University of Toronto\\
{\tt\small weicheng.tseng@mail.utoronto.ca}
\and
Fu-En Wang\\
National Tsing Hua University\\
{\tt\small secondauthor@i2.org}
\and
Min Sun\\
National Tsing Hua University\\
{\tt\small secondauthor@i2.org}
}
\fi
\author{
    Yi-Shan Lee$^{1}$ \\
    {\tt\small yishanlee@m110.nthu.edu.tw}
    %{\tt\small Fuen\_Wang@asus.com}
    \and
    Wei-Cheng Tseng$^{2}$\\
    {\tt\small weicheng.tseng@mail.utoronto.ca}
    \and
    Fu-En Wang$^{1}$\\
    {\tt\small fulton84717@gapp.nthu.edu.tw}
    \and
    Min Sun$^{1}$\\
    {\tt\small sunmin@ee.nthu.edu.tw}
}

\maketitle
\ificcvfinal\thispagestyle{empty}\fi

%------------------------------------------------------------------------
\myfootnote{1}{National Tsing Hua University}
\myfootnote{2}{University of Toronto}
% \myfootnote{3}{NEC Labs America}
% \myfootnote{4}{\scriptsize MOST Joint Research Center for AI Technology and All Vista Healthcare}

\def\faceloss{cross-modality lifting loss}
\def\simloss{cross-modality similarity loss}
\def\frameworkname{VMCML}
\def\opensetname{unseen music}
\def\closesetname{seen music}
\def\ourdataset{MSVD}
\def\opensettable{Unseen Music}
\def\closesettable{Seen Music}
\def\music{music}

\begin{abstract}
    We propose a content-based system for matching video and background \music. The system aims to address the challenges in music recommendation for new users or new music give short-form videos. 
    %is the measurement of the similarity between video and music for new users and/or new musics.
    %since the instability from cross-modal inputs usually leads to a sub-optimal training results. 
    To this end, we propose a cross-modal framework \frameworkname{} that finds a shared embedding space between video and music representations. To ensure the embedding space can be effectively shared by both representations, we leverage CosFace loss based on margin-based cosine similarity loss.
    Furthermore, we establish a large-scale dataset called \textbf{\ourdataset{}}, in which we provide 390 individual \music{} and the corresponding matched 150,000 videos.
    %short video background music dataset(SVBMD)\fuen{WTF}
    %
    % Furthermore, we establish a large-scale content-based video background music recommendation dataset, datasetofname, composed of approximately 390 different music clips associated with 150,000 videos.  QAQ
    We conduct extensive experiments on \textbf{Youtube-8M} and our \textbf{\ourdataset{}} datasets. Our quantitative and qualitative results demonstrate the effectiveness of our proposed framework and achieve state-of-the-art video and music matching performance.
    %We provide both quantitative and qualitative results on the established short video background music dataset(SVBMD) to show the effectiveness of the proposed approach.
    % Experiments on the established datasetofname demonstrate the effectiveness of the proposed method. A qualitative recommendation result is also included.
\end{abstract}
\section{Introduction}
In recent years, short-form videos have rapidly entered our daily lives. People record their life by uploading short videos to various platforms such as TikTok, Instagram Reels, and Youtube Shorts. 
The daily usage time of TikTok by young generation users (ages 
10-25) in 2022 has grown by 2.38 times from 2017 to 2022, with 44 minutes into 105 minutes of daily TikTok. After the opening of Instagram Reels in April 2022, the number of short-form videos has increased significantly with an increase of 971 percent in August compared to April.
These platforms typically provide context-based recommendation systems to help users attach background \music{} to their uploaded videos. These \music{} are mostly selected according to users' previous selection or current trends, which eventually bias towards a few existing trendy background music.
The diversity of background music is important for these platforms since new trendy music can be more engaging and the uniqueness of trendy music can attract more new users to the platform.
However, the contents of background music and videos are typically not considered.
\begin{figure}[t]
    \begin{center}
        \includegraphics[width=1.0\linewidth]{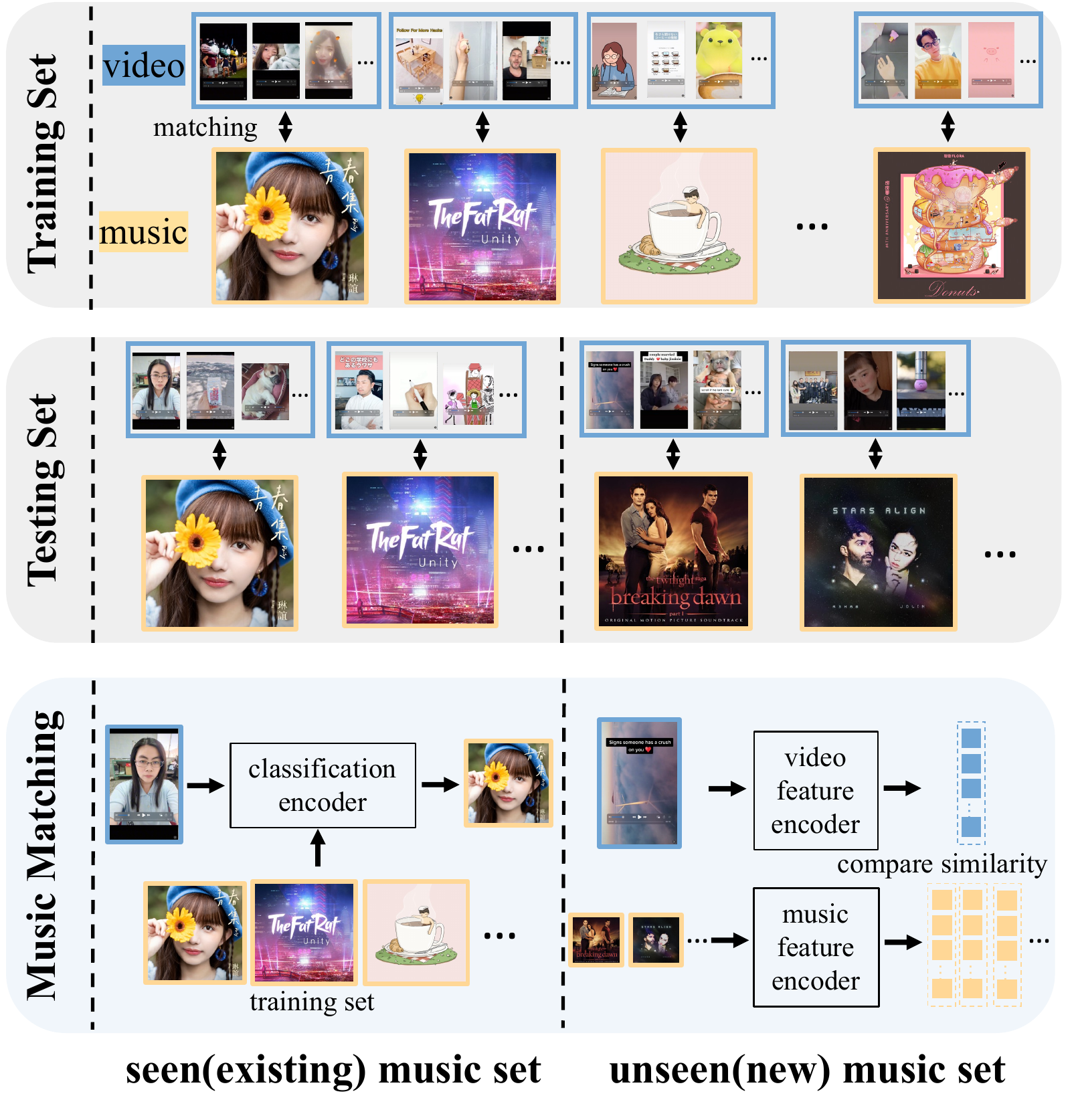}
    \end{center}
    \caption{\textbf{The content-based video and music matching system.} For training, we collect several pieces of music and several videos using the same background music. Hence, one music matches several videos. 
    In testing, \closesetname{} set (existing music in training) can be addressed as classifying a video to a set of music identities. The \opensetname{} set (new music unseen in training) can be viewed as performing video verification by comparing the similarity between the video and the new music identities.}
    %We define the \closesetname{} set of testing as the music that exists in the training set; the \opensetname{} set of testing as the music that is new from the training set. 
    % \fuen{You need a teaser.} [Wei-Cheng] I agree.}
    \label{fig:teaser}
\end{figure}

The content-based music matching system, as shown in Figure~\ref{fig:teaser}, is crucial for 
(1) recommending new music as they haven't been selected and (2) recommending existing music to new users as they lack of previously selected music.  
There are few pioneers who work on cross-modal matching for video and audio content.
Hong~\etal~\cite{hong2018cbvmr} proposed a content-based retrieval model that combines the inter-modal ranking loss and soft intra-modal structure loss to construct a shared embedding space.
Sur{\'\i}s~\etal~\cite{suris2018cross} (referred to as CEVAR) proposed a joint embedding model with the classification loss along with a similarity loss, which incorporates the video labels provided by the YouTube-8M~\cite{abu2016youtube} dataset.
Yi~\etal~\cite{yi2021cross} (referred to as CMVAE) proposed a hierarchical Bayesian generative model using variational auto-encoder and matching relevant background music to videos by the corresponding latent embeddings.
However, all these methods suffer from a big performance gap between matching existing music to new videos v.s. matching new music to new videos. We use seen or existing interchangeably and unseen or new interchangeably.

%The content-based music matching system is crucial for attracting new multimedia platform users and increasing the traffic of new musics. 
%The previous behavior information of new users is insufficient, and new music lacks user usage data. 
%There are two major challenges for video-music matching systems. 1) The previous behavior information of new users is unknown and insufficient. 2) The new musics are lack of enough usage data for being precisely utilized into matching systems.
%To solve the above challenges, 
We treat this cross-modal matching task as a metric-learning problem where most \music{} matches to a few videos as they are non-trendy. Moreover, we want our method improves on both recommending existing music, as well as new music.
%We design a cross-modal content-based system ``\textbf{XXXX}''. As illustrated in Figure~\ref{fig:teaser}, our proposed framework can be generalized to both trained and unseen videos and musics. For unseen musics, the video-music matching task can be formulated into a metric-learning problem by calculating the feature similarity between videos and musics For trained musics, we consider the task as a classification problem since the music prototypes are included in the final classification layer of the proposed network.
%\ys{In testing, matching music can be evaluated under trained music set and unseen music set. For the trained music set, all testing musics are in the training set. 
%Therefore, the videos in the trained music set can be solved as a classification problem. For the unseen music set, the testing musics are out of the training set, making music matching video more challenging. It would be overhead if we retrained whenever we had new music. 
%We hope that when there is new music, the previously trained model can be directly applied. Since it is incredible to classify videos by the musics in training set, we need to calculate the similarity between the video representation and music representation.
%Furthermore, \ys{Most non-trending music is pair less with videos, and the ground truth number of the background music matching for video is only one. 
%The ground truth comes from the music chosen by the uploader.}
It is not obvious at first glance, but face recognition~\cite{turk1991face,sun2014deep,liu2017sphereface,kemelmacher2016megaface} shares the same challenges to our task. A face recognition model not only needs to recognize existing faces in training. It also needs to enroll new faces and identify them without retraining the model.
In addition, the model should only require a person to enroll a few faces for ease of usage.
Inspired by these observations, we adopt the CosFace loss~\cite{cosface} and the ArcFace loss~\cite{deng2019arcface}, widely used for face recognition to our cross-modal matching task. We refer to these losses as lifting loss. To find a shared embedding space between video and music, there is a shared head that makes both video and music features more aligned. Furthermore, we calculate the similarity loss between video and music features. More specifically, we combine \faceloss{} and \simloss{} for cross-modal matching. 

There is no publicly available dataset for matching video and background music. 
% for us to \ys{conduct} experiments. 
% It is challenging to construct such a dataset since the pairs of video and music on these platforms are unorganized and constantly changing. In order to make sure the quality, we discard most of them and only leave the videos with official background music. 
We established a dataset, Music for Short Video Dataset \textbf{\ourdataset{}}, containing 150k videos and 390 corresponding background \music{} pairs. The differences between Youtube-8M~\cite{abu2016youtube} and MSVD are (1) the ground truth of music and video in the former is a one-to-one mapping, (2) most of the music in the former is the soundtrack of the video and (3) most of the videos in the former are longer than 1 minute.
We divide the MSVD into \closesetname{} set and \opensetname{} set corresponding to real-world scenarios.  
% We extract video features and music features using generic networks pre-trained on large datasets.  
With the \textbf{\ourdataset{}} dataset established, we conduct experiments to evaluate the performance of \frameworkname{} and compare it with previous methods.

Our main contributions are summarized as follows:
\begin{itemize}
    \item[(1)] We propose a novel cross-modal framework \frameworkname{} with \faceloss{} and \simloss{} for content-based background music matching, which can be applied to both seen and unseen video-music samples.
    
    \item[(2)] We collect a short video and background music matching dataset called ``Music for Short Video Dataset~(\ourdataset{})'', the first publicly available dataset for the video-music matching task which provides 390 \music{} and their 150,000 corresponding videos.

    \item[(3)] \frameworkname{} achieve the state-of-the-art on \textbf{\ourdataset{}} and Youtube-8M~\cite{abu2016youtube}.
\end{itemize}
\section{Related Works}
We introduce the previous cross-modal matching methods in Sec.~\ref{cross_modal_matching}, and present the metric learning in Sec.~\ref{metric_learning}. 
\subsection{Cross-Modal Matching}\label{cross_modal_matching}
Most existing cross-modal matching methods typically focus on textual and visual modalities, using open-sourced datasets such as MSCOCO~\cite{lin2014microsoft}, Flickr30k~\cite{plummer2015flickr30k}, ActivityNet-captions~\cite{krishna2017dense} and MSR-VTT~\cite{xu2016msr}.
Zhen~\etal~\cite{zhen2019deep} proposed a deep supervised cross-modal retrieval method that simultaneously minimizes discrimination loss in the label space and invariance loss.
Wei~\etal~\cite{wei2020universal} introduced a universal weighting metric learning framework and a new polynomial loss under it. The universal weighting framework provides a powerful tool to analyze various metric-learning-based weighting and loss, which have been widely used in cross-modal matching.
However, there has been little effort devoted to matching video and music content. Hong~\etal~\cite{hong2018cbvmr} presented a content-based retrieval model that only uses content features between music and videos, constraining the relative distance relationship of samples within each modality. While Suris~\etal~\cite{suris2018cross} uses the visual features and audio features provided by Youtube-8M~\cite{abu2016youtube} to minimize the distance of visual embedding and audio embedding of the same video at representation space to predict the corresponding video label.
Yi~\etal~\cite{yi2021cross} proposed a hierarchical Bayesian generation model using the variational auto-encoder to match relevant background music to video through their latent embedding.
Nonetheless, these methods face a significant performance gap when matching existing music to new videos compared to matching new music to new videos. This work introduces face recognition-inspired loss to mitigate the gap.

\begin{figure*}[t]
    \begin{center}
        \includegraphics[width=1.0\linewidth]{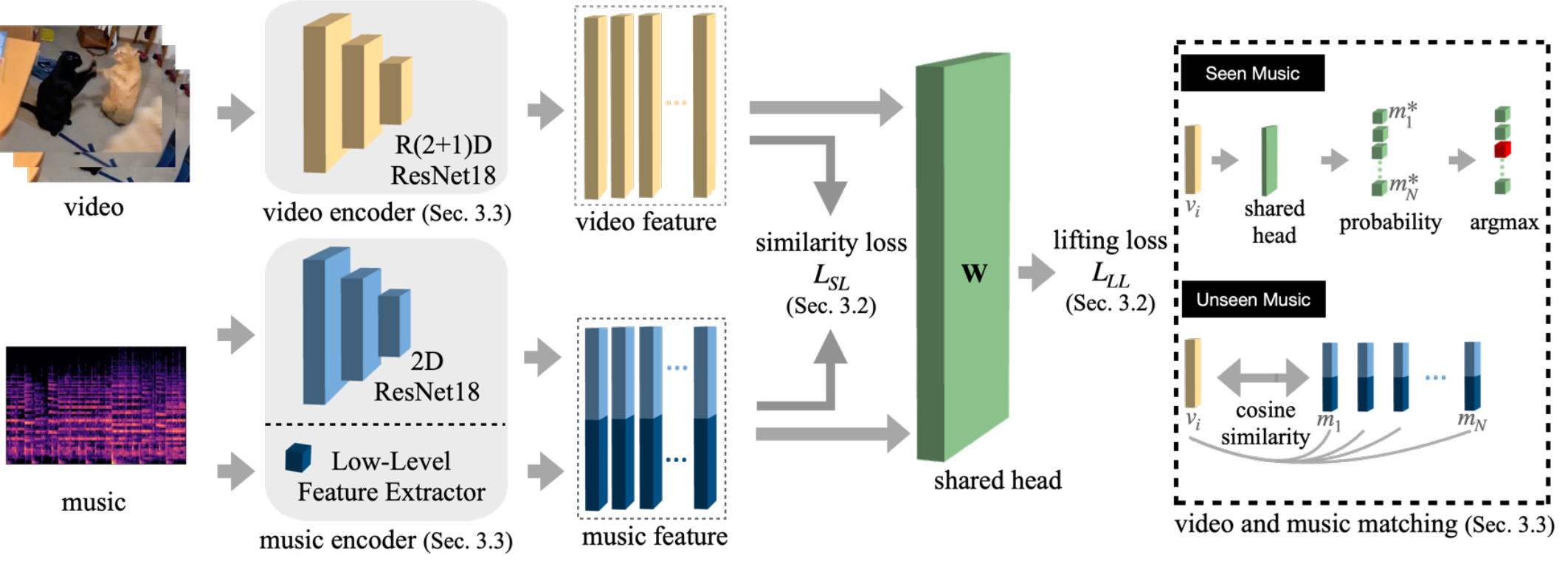}
    \end{center}
    \caption{\textbf{Overview of the proposed \frameworkname~framework} for video-music matching. Given a pair of video and music, the video encoder is applied to extract video features and the music encoder is adopted to collect music features(Sec.~\ref{m_e}). There are two parts of the music encoder, and we concatenate their output as music features. In the training step, we calculate the \faceloss{} for the video and music separately from video features and music features by utilizing the same shared head (Sec.~\ref{loss_class}). Hence, calculate the similarity loss between video features and music features(Sec.~\ref{loss_sim}). In the testing step (the right side of the figure), we treat the task as a classification problem on seen music set and calculate the cosine similarity between video and music on unseen music set to match video with appropriate music (Sec.~\ref{matching}).}
    \label{fig:framework}
\end{figure*}

\subsection{Metric learning}\label{metric_learning}
The goal of metric learning is to learn the similarity between features to enable accurate feature matching and verification. To improve the quality of feature embedding, contrastive loss~\cite{chopra2005learning,deng2009imagenet} and triplet loss~\cite{wang2014learning,hoffer2015deep} are commonly employed techniques that help increase the Euclidean margin. In addition, there are other variations of metric learning methods such as center loss~\cite{wen2016discriminative} and angular loss~\cite{cosface} that have been proposed specifically for face recognition tasks. Center loss aims to minimize the distance between each sample and its corresponding class center. In contrast, angular loss minimizes the distance between the feature embedding and its corresponding class boundary. Overall, metric learning has played a crucial role in enabling accurate verification systems for learning the similarity between features.

\section{Approach}

% In this section, we first define our video music pair in Section~\ref{pro_def}. 
In this section, we introduce the problem formulation of the video-music matching task in Sec.~\ref{pro_def}. Then, we detail our loss functions in Sec.~\ref{objective} for lifting cross-modality features to a shared space.
Finally, we introduce our proposed \frameworkname~framework for video-music matching in Sec.~\ref{v_e}.

\subsection{Problem Definition}\label{pro_def}
Given a music set $\mathcal{M}=\{m\}_{i=1}^{N_m}$ with $N_m$ music and a video set $\mathcal{V}=\{v\}_{i=1}^{N_v}$ with $N_v$ videos from training dataset, where $(m,v)$ denote \music~and video samples. Our video-music matching task is formulated as two transformation functions ${\rm f}(m) \rightarrow y_m$ and ${\rm g}(v) \rightarrow y_v$, and $(y_m,y_v) \in [1,N_m]$ denote the predicted matching \music{} indexes in the training dataset. For matching video and \music{} as a metric learning problem, we further adopt the  shared weight $\textbf{W}$ to lift video and music features to a shared embedding space: 
\begin{align}
    \begin{split}
        % &{\rm h}({\rm f}(m)) \rightarrow y_m~,\\
        &\mathbf{W}\cdot{\rm f}(m) \rightarrow y_m~,\\
        &\mathbf{W} \cdot g(v) \rightarrow y_v~.
    \end{split}
    \label{eq:f-g}
\end{align}
During the testing stage, video and \music{} can be matched by estimating the cosine similarity between their features, i.e., ${\rm cos}({\rm f}(m),{\rm g}(v))$. For \music{} already included in the training dataset, we can directly use $y_v$ as the matched one. In this paper, the transformation functions $\rm f$ and $\rm g$ are respectively referred to as a video and a music encoder of which output embedding dimensions are both chosen as $l$.

\subsection{Cross-Modality Training Objectives}\label{objective}
Softmax loss is commonly used in the classification problem for minimizing intra-class and maximizing inter-class distances, which is formulated as:
\begin{equation}
    \begin{aligned}
        L_S(x_i, \mathbf{W}) = -\log \frac{e^{\mathbf{W}_{y_i} \cdot x_i}}{\sum_{j=1}^N e^{\mathbf{W}_j \cdot x_i}}~,
    \end{aligned}
    \label{eq:softmax-loss}
\end{equation}
where $\mathbf{W}$ is the prototype of each class, i.e., the weight of the last layer in a network, $N$ is the total number of classes, $x_i$ is the feature, and $y_i$ is the ground truth class index of $x_i$. For further improving the decision boundary between different classes, CosFace~\cite{cosface} proposed to lift the features and prototypes to a hyper-sphere by introducing a scaling term $s$ and a margin $\mu$:

\begin{equation}
\begin{aligned}
    &\cos(\theta_{k, i}) = \frac{\mathbf{W}_{k} \cdot x_i}{\norm{\mathbf{W}_{k}} \cdot \norm{x_i}}~, \\
    L_{C}(x_i, \mathbf{W})=&-\log \frac{e^{s \cdot [\cos(\theta_{y_i,i})-\mu]}}{e^{s \cdot [\cos(\theta_{y_i,i})-\mu]} + \sum_{j \neq i}^N e^{s \cdot \cos(\theta_{j,i})}}~.
    \label{eq:cosface}
\end{aligned}
\end{equation}
Since Equation~\eqref{eq:cosface} is based on the angles between intra and inter classes (i.e., $\theta_{y_i,i}$ and $\theta_{j,i}$) in a normalized feature space, the features $x_i$ are thus optimized in a hyper-sphere.

\paragraph{Cross-Modality Lifting Loss.}
\label{loss_class}
\fuen{
To solve the video-music matching task as a metric learning problem, we aim at lifting video and music features to the same hyper-sphere. In this way, we can match the videos to their most appropriate music by calculating the cosine similarity between them. Hence, we propose ``Cross-Modality Lifting Loss'' by adopting a shared prototype $\mathbf{W}$ for both video and music features and considering modality-to-prototype distances:
\begin{equation}
\begin{aligned}
    L_{LL} ({\rm g}(v), {\rm f}(m), \mathbf{W}) = L_C({\rm g}(v), \mathbf{W}) + \alpha L_C({\rm f}(m), \mathbf{W})~,
    \label{eq:LL}
\end{aligned}
\end{equation}
where $(v,m)$ are input music and video, $(\rm g,f)$ are the transformation functions as described in Equation~\eqref{eq:f-g}, and $\alpha$ is a hyper-parameter. In practice, the transformation functions are implemented as two independent feature encoders for video and music.}

\paragraph{Cross-Modality Similarity Loss.}
\label{loss_sim}
\begin{table*}[ht]  
\begin{center}   
\resizebox{\textwidth}{!}{\begin{tabular}{l||r||cc|cc|cc|
cc}  
  
\toprule % Toprule applied here  
  
\textbf{Dataset}&\multicolumn{1}{c||}{} & \multicolumn{2}{c|}{\textbf{Training}} & \multicolumn{2}{c|}{\textbf{Validation}} & \multicolumn{2}{c|}{\textbf{\closesettable}} & \multicolumn{2}{c}{\textbf{\opensettable}}\\
% & \multicolumn{2}{c|}{} & \multicolumn{2}{c|}{} & \multicolumn{2}{c|}{\textbf{(Weak generalization)}} & \multicolumn{2}{c|}{\textbf{(Strong generalization)}} &\\
% 
&\#Video &\#Music & VpM  & \#Music & VpM  & \#Music & VpM  & \#Music & VpM \\  
\midrule % Midrule applied here  
 \ourdataset{} & 150000 &265 & 400 & 265 & 20 & 265 & 80 & 125 & 140\\ 
 \midrule % Midrule applied here 
 Youtube-8M~\cite{abu2016youtube} & 5678 & 4654 & 1 & NA & NA & NA & NA & 1024 & 1024\\   
\bottomrule % Bottomrule applied here  
\end{tabular}}%
\caption{\textbf{Details of datasets.} The music for training, validation and \closesetname{} set is the same. Because the video-music pair is one-to-one mapping in Youtube-8M~\cite{abu2016youtube} dataset, validation and \closesetname{} are not included. \textbf{VpM} indicates the number of matching videos for each music.}  
\label{tab:table_msvd} 
\end{center}  
\end{table*}  
\fuen{
Although our proposed \faceloss~can effectively minimize the intra and maximize the inter class distances, we found that only considering such a modality-to-prototype distance still leads to a sub-optimal performance since videos and music are eventually matched based on their features instead of their prototype during the testing stage. To this end, we follow \cite{suris2018cross} to adopt ``Cross-Modality Similarity Loss'' aiming at addressing the video-to-music feature distances for improving our downstream video-music matching performance:
\begin{equation}
\begin{aligned}
    \cos(x_i, x_j) &= \frac{x_i \cdot x_j}{\norm{x_i} \cdot \norm{x_j}}~, \\
    L_{SL} ({\rm g}(v), {\rm f}(m), {\rm f}(m^\prime)) &= max[\tau, \cos({\rm g}(v), {\rm f}(m^\prime))] \\
    &- \cos({\rm g}(v), {\rm f}(m))~,
    \label{eq:SL}
\end{aligned}
\end{equation}
where $(v, m)$ indicates a positive video-music pair queried according to the ground truth music index of $v$, $(v, m^\prime)$ indicates a negative one randomly sampled from the dataset, and $\tau$ is a selected margin value. With \simloss, we can apply direct constraints in-between the predicted video and music features, and consider modality-to-modality distances, which provides consistent video-music matching schemes under both training and testing stages.
}

\subsection{\frameworkname~Framework}
\paragraph{Video Encoder.}
\label{v_e}
\fuen{To extract video features, we adopt a R(2+1)D ResNet-18~\cite{tran2018closer}, pretrained on Kinetics-400~\cite{kay2017kinetics} and followed by a fully-connected layer to infer the video features with embedding size $l$, as our video encoder ($\rm g$ in Equation~\eqref{eq:f-g}).}

\paragraph{Music Encoder.}
\label{m_e}
\fuen{we firstly calculate the Mel spectrograms of the input \music{} and adopt a ResNet-18~\cite{he2016deep} pretrained on ImageNet~\cite{deng2009imagenet} to extract the high-level music features. In addition, inspired by \cite{yi2021cross}, we use openSMILE~\cite{eyben2010opensmile} to extract the low-level music features, including MFCC, voice intensity, pitch, etc. The low-level and high-level features are then fused by concatenation and we infer the final music features by passing the fused features into an additional fully-connected layer with embedding size $l$. We refer this music encoder as $\rm f$ in Equation~\eqref{eq:f-g}.}

\paragraph{Training and Testing.}
\label{matching}
\fuen{During each training iteration, we use a pair of video and music for calculating $L_{LL}$ (Equation~\eqref{eq:LL}). For $L_{SL}$ (Equation~\eqref{eq:SL}), we randomly sample a negative music for calculation. The final training objective is established as:
\begin{equation}
\begin{aligned}
L(v, m, m^\prime) = L_{LL}({\rm g}(v), f(m), \mathbf{W})& + \beta L_{SL}({\rm g}(v), {\rm f}(m),\\
{\rm f}(m^\prime))
% &+ L_{SL}({\rm g}(v), {\rm f}(m^\prime))]~,
\label{eq:L-final}
\end{aligned}
\end{equation}
where $m^\prime$ indicates a randomly-sampled negative music, and $\beta$ is a hyper-parameter.}

\fuen{During the testing stage, the \closesetname~and \opensetname~sets are evaluated with different schemes, as illustrated in the right part of Fig.~\ref{fig:framework}.
For \closesetname~set, the video features are inferred from our video encoder $\rm g$, while the music features are directly pulled from the trained prototype since the trained prototype can represent the feature center of each training music, which can significantly reduce the inference time of \opensetname~features. 
For \opensetname~set, the music features are extracted from our music encoder $\rm f$.}

\fuen{Finally, we match the videos and music by calculating the cosine similarity between their features and select the top 20 music clips with the highest similarities as the matched music list.}
\section{Experiments}
% In this section, we first describe the datasets in Section~\ref{Dataset}.
% To verify the effectiveness of the proposed method \frameworkname{}, we conduct experiments on MSVD and a widely-used benchmark dataset: Youtube-8M~\cite{abu2016youtube}. The implementation details are shown in Section~\ref{imple} and the introduction of the evaluation metric we use in our experiment is in Section~\ref{metric}. In Section~\ref{eva_MSVD}, we compare the proposed \frameworkname{} method with state-of-the-art techniques to evaluate its performance on existing and new music. Assessing them on another cross-modal dataset, Youtube-8M~\cite{abu2016youtube}, shown in Section~\ref{eva_yt8m}. We provide our result on quantitative and qualitative in the following.

% \wc{
In this section, we describe the datasets including \ourdataset{} and Youtube-8M~\cite{abu2016youtube} in Sec.~\ref{Dataset}. The implementation details and evaluation metric are presented in Sec.~\ref{imple}. Moreover, we provide quantitative and qualitative results for the comparison between our approach and state-of-the-art techniques in Sec.~\ref{eva_MSVD}. The ablation study is in Sec.~\ref{ablation}.
% }

\subsection{Dataset}\label{Dataset}
% Table~\ref{tab:table_msvd} shows details of the datasets employed in our experiments.

% \wc{
We summarize the details about the datasets in Table~\ref{tab:table_msvd} and further illustrate them as follows.
% }
\paragraph{\ourdataset{}.}
Matching short videos with background music requires a suitable dataset, but there is no publicly available one~\footnote{
The dataset used in CMVAE is not publicly available.
% For the CMVAE, they don't release dataset.
}. The Youtube-8M~\cite{abu2016youtube} dataset contains video features and music features, but most \music{} are the soundtrack of videos. Additionally, most videos are longer than one minute. To overcome these limitations, we collect videos with background music from a well-known multimedia platform. We filter out videos whose music (1) is the soundtrack of the video or (2) upload by the video uploaders. Furthermore, we exclude those videos longer than 20 seconds. We download corresponding videos randomly for each \music{}. 
The video resolution is down-sampled by one-fourth to reduce data size. Our Music for Short Video Dataset (\textbf{\ourdataset{}}) comprises about 150,000 video-music samples, each 8 seconds long and one frame per second. We used the uploader's choice of background \music{} as the ground truth for each sample. In summary, the dataset includes 390 \music{}, split into a \closesetname{} set (265 clips) and an unseen music set (125 clips). The \closesetname{} set was randomly divided into training, validation, and testing sets, with a ratio of 20:1:4 for 500 corresponding videos. The \opensetname{} set contains 140 videos for each \music{}, with no music present in the training music set.
%Unofficial music is often made by users and is at risk of not being generalized well by other videos.

\paragraph{Youtube-8M~\cite{abu2016youtube}.}
Youtube-8M~\cite{abu2016youtube} is not ideal for our proposed use case since the majority of the music in video-music pairs is the natural sound from the video. 
We follow CEVAR~\cite{suris2018cross} to conduct experiments on a random subset of 6000 clips. We use the pre-computed video-level features in the dataset: a single vector for audio information and a single vector for visual information in a video. All experiments use these fixed pre-computed features.
Youtube-8M includes video genre classification labels that indicate the topic of the video clip, and CEVAR utilized the labels as additional training signals.
We ignore the video genre classification labels to make the experiment setting the same as our MSVD dataset.

\begin{table*}[th]  
\begin{center}   
\resizebox{\textwidth}{!}{\begin{tabular}{l|cccc|cccc}  
  
\toprule % Toprule applied here  
  
&\multicolumn{4}{c|}{\textbf{\closesettable}} & \multicolumn{4}{c}{\textbf{\opensettable}}\\

& Recall@1 & Recall@5 & Recall@10 & Recall@20 & Recall@1 & Recall@5 &Recall@10 & Recall@20 \\  
  
\midrule % Midrule applied here  
   
 \textbf{\frameworkname{}$_\text{Cosface}$} & \textbf{0.2056} & \textbf{0.3562} & \textbf{0.4409} & \textbf{0.5450} & \textbf{0.0170} & \textbf{0.0678} & \textbf{0.1329} & \textbf{0.2526}\\
\frameworkname{}$_\text{Arcface}$ & 0.1685 & 0.2772 & 0.3370 & 0.4203 & 0.0145 & 0.0673 & 0.1298 & 0.2455\\ 
 \midrule % Midrule applied here 
 Random & 0.0037 & 0.0188 & 0.0377 & 0.0754 & 0.0080 & 0.0400 & 0.0800 & 0.1600\\
 % \midrule % Midrule applied here 
 CEVAR~\cite{suris2018cross} & 0.1883 & 0.3319 & 0.4188 & 0.5300 & 0.0156 & 0.0613 & 0.1153 & 0.2170\\  
 CMVAE~\cite{yi2021cross} & 0.0277 & 0.0817 & 0.1316 & 0.2040 & 0.0112 & 0.0474 & 0.0906 & 0.1740\\  
  
\bottomrule % Bottomrule applied here  

\end{tabular}}%
\caption{Comparison the performance (Recall@K) between the proposed \frameworkname{} with baseline methods on \ourdataset. \frameworkname{}$_\text{Cosface}$ indicates that the \faceloss{} of the framework is CosFace loss~\cite{cosface}; \frameworkname{}$_\text{Arcface}$ represents that the \faceloss{} of the framework is ArcFace loss~\cite{deng2019arcface}. All architectures we implement use the same video encoder and music encoder.} 
\label{tab:method} 
\end{center}  
\end{table*}

% \begin{table}[h]
%     \begin{subtable}[h]{0.45\textwidth}
%        % \caption{\closesettable}
%         \centering
%         \resizebox{\columnwidth}{!}{\begin{tabular}{l|cccc}  
          
%         \toprule % Toprule applied here  
%         & \multicolumn{4}{c}{\closesettable}\\
%         & Recall@1 & Recall@5 & Recall@10 & Recall@20\\  
          
%         \midrule % Midrule applied here  
           
%         \textbf{Resnet18} & \textbf{0.2056} & \textbf{0.3562} & \textbf{0.4409} & \textbf{0.5450}\\  
        
%         Vggish & 0.1999 & 0.3453 & 0.4325 & 0.5358\\   
          
%         \bottomrule % Bottomrule applied here  
          
%         \end{tabular}}% 

%        \label{tab:mbk_e}
%     \end{subtable}
%     \hfill
%     \vspace{10pt}
%     \begin{subtable}[h]{0.45\textwidth}
%         % \caption{\opensettable}
%         \centering
%         \resizebox{\columnwidth}{!}{\begin{tabular}{l|cccc}  
          
%         \toprule % Toprule applied here  
%         & \multicolumn{4}{c}{\opensettable}\\
%         & Recall@1 & Recall@5 & Recall@10 & Recall@20\\  
          
%         \midrule % Midrule applied here  
           
%         \textbf{Resnet18} & \textbf{0.0170} & \textbf{0.0678} & \textbf{0.1329} & \textbf{0.2526}\\  
        
%         Vggish & 0.0121 & 0.0574 & 0.1129 & 0.2155\\  
          
%         \bottomrule % Bottomrule applied here  
          
%         \end{tabular}}%
%         \label{tab:mbk_n}
%      \end{subtable}
%      \caption{Comparison of Resnet18 and Vggish for music encoder backbone on \frameworkname{} architecture. Vggish is the backbone originally used by CMVAE as the music encoder.}
%      \label{tab:mbk}
% \end{table}

\begin{table*}[th]
    \centering
    \resizebox{\textwidth}{!}{\begin{tabular}{l|cccc|cccc}  
          
    \toprule % Toprule applied here  
    & \multicolumn{4}{c|}{\closesettable} & \multicolumn{4}{c}{\opensettable}\\
    & Recall@1 & Recall@5 & Recall@10 & Recall@20 & Recall@1 & Recall@5 & Recall@10 & Recall@20\\  
          
    \midrule % Midrule applied here  
           
    \textbf{ResNet-18} & \textbf{0.2056} & \textbf{0.3562} & \textbf{0.4409} & \textbf{0.5450} & \textbf{0.0170} & \textbf{0.0678} & \textbf{0.1329} & \textbf{0.2526}\\  
        
    Vggish & 0.1999 & 0.3453 & 0.4325 & 0.5358 & 0.0121 & 0.0574 & 0.1129 & 0.2155\\   
          
    \bottomrule % Bottomrule applied here  
          
    \end{tabular}}% 
    \caption{Comparison of Resnet18 and Vggish for music encoder backbone on \frameworkname{} architecture. Vggish is the backbone originally used by CMVAE as the music encoder.}
     \label{tab:mbk}
\end{table*}
\subsection{Implementation Details}\label{imple}
\fuen{
We implement our proposed framework using three NVIDIA RTX 3090 with PyTorch~\cite{paszke2019pytorch}. All of the network parameters, including video encoder, music encoder, and shared prototype, are jointly optimized by Equation~\eqref{eq:L-final}. 
Adam~\cite{kingma2014adam} optimizer is adopted with learning rate 1e-5, and both the margin values $\mu$ and $\tau$ are set to $0.2$. The weight decay is set to $0.002$, and the batch size is set to $128$ for all models. The embedding size $l$ of video and music features is set to 256. Hyper-parameters are selected based on the evaluation metric on the validation set at recall@10. $\alpha$ and $\beta$ in Equation~\eqref{eq:LL} and \eqref{eq:L-final} are $0.38$ and $2$, respectively.}

\paragraph{Evaluation Metric.}\label{metric}
The evaluation metric Recall@K is denoted as:
\begin{equation}
    \text{Recall}@K=\sum_{v\in \mathcal{V}^{te}}\dot\#Hits_v@K
\end{equation}
where $\mathcal{V}^{te}$ denotes the set of testing video.
The Recall@K indicates the percentage of queries for which the model returns the correct item in its top K result.

% \subsection{Comparison of methods}
\subsection{Experimental Results}
\paragraph{Baselines.}
\fuen{To demonstrate the effectiveness of the proposed framework, we compare our \frameworkname~with the state-of-the-art video-music matching approaches. 1) CEVAR~\cite{suris2018cross}: the approach adopting softmax loss for video-music matching. 2) CMVAE~\cite{yi2021cross}: the approach adopting a variational auto-encoder for video-music matching.}

%which are based on supervised and variational auto-encoder methods, respectively. For CEVAR, the shared head is a fully connected layer and uses softmax loss for classification. For CMVAE, the loss for video and music is calculated using a cross-modal variational auto-encoder. We also provide the random guess result as a reference, which is $k/n$, where $k$ represents the top $k$ and $n$ represents the number of videos in the testing set.

\paragraph{Comparison on \ourdataset.}
\label{eva_MSVD}
\fuen{The quantitative results on our proposed \ourdataset~ dataset is shown in Table~\ref{tab:method}. In general, our proposed method outperforms the other baselines in all metrics. Compared to CMVAE, we found that learning the distribution of videos and music with a variational auto-encoder might not be a satisfying solution for video-music matching task, and our method improved Recall@10 by $2.4\%$ on \closesetname~and $46.7\%$ on \opensetname~. On the other hand, although CEVAR also learns a shared prototype with a softmax loss function, we improve Recall@10 by $5.3\%$ on \closesetname~and $15.3\%$ on \opensetname. We found that lifting video and music features to a hyper-sphere by our \faceloss~ is necessary for improving the decision boundary between classes.}

%our method improved Recall@10 by $5.3\%$ on \closesetname{} and $15.3\%$ on \opensetname{}. Softmax loss emphasizes the correct classification while ignoring the discernment between other features except for the correct class. However, the \faceloss{} emphasizes the discriminative from features. Compared to CMVAE, a variational-based generative, our method improved Recall@10 by $2.4\%$ on \closesetname{} and $46.7\%$ on \opensetname{}. 

\begin{figure}[h]
    \begin{center}
        \includegraphics[width=1.0\linewidth]{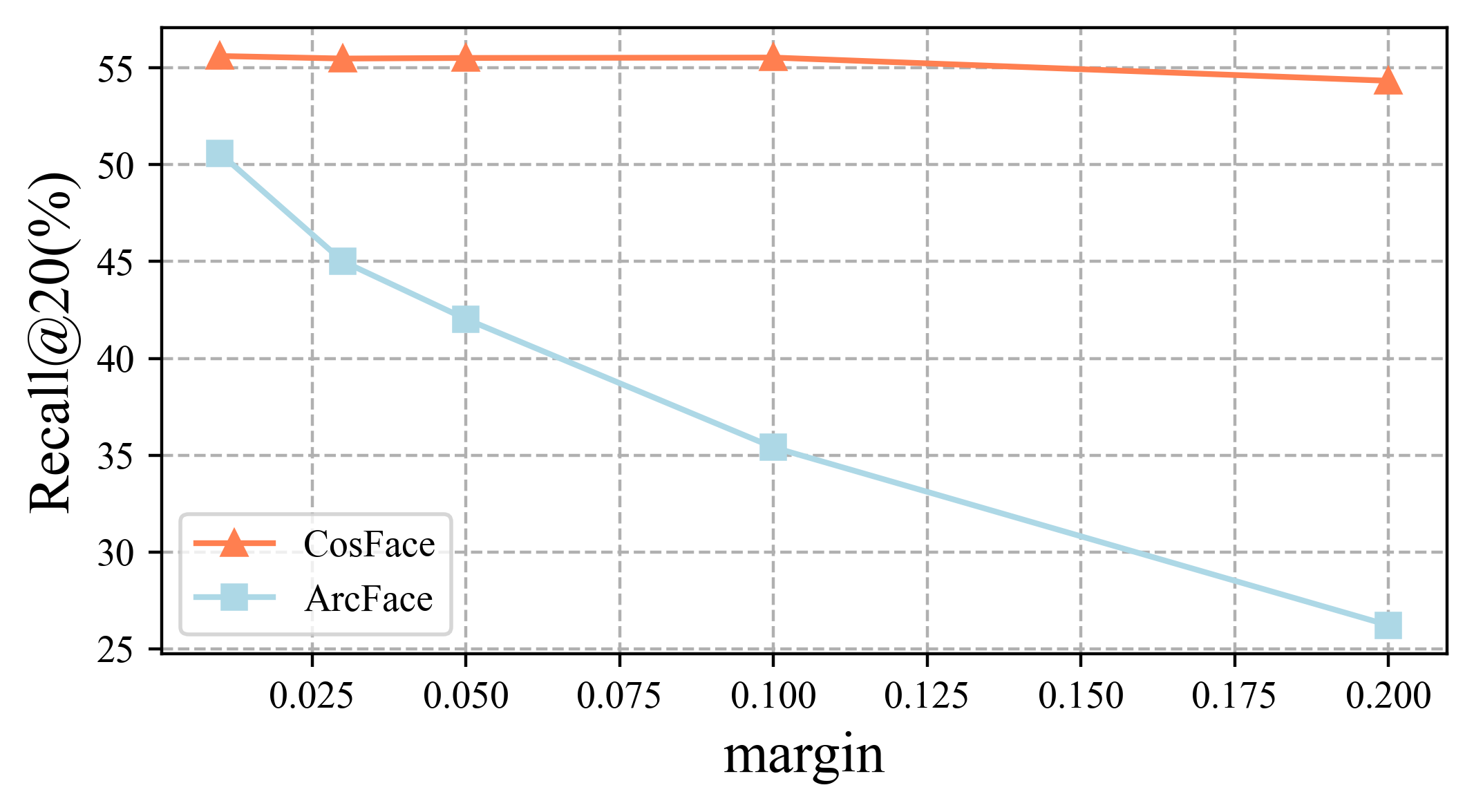}
    \end{center}
    \caption{Recall@20(\%) with different margin parameters $\mu$ on CosFace~\cite{cosface} and ArcFace~\cite{deng2019arcface}. We implement the experiments with \frameworkname{} on \closesetname{} set.}
    \label{fig:margin}
\end{figure}

\fuen{In addition to the comparison with SOTA approaches, we also compare the performance difference between CosFace and ArcFace loss functions. We empirically found that CosFace leads to a better training convergence, and thus we adopt CosFace to lift our video and music features to a shared hyper-sphere. Moreover, we conduct experiments for selecting the best margin value $\mu$ in Equation~\eqref{eq:LL} as shown in Fig.~\ref{fig:margin}. We evaluate the recall@20 performance on the \closesetname~set of MSVD by increasing $\mu$ from $0.01$ to $0.2$. The results indicate that CosFace achieves better performance and training stability than ArcFace does. As $\mu$ increases from $0.1$ to $0.2$, the variation of the ArcFace performance is approximately 0.1, while the variation of CosFace performance is smaller than 0.01. These findings suggest us to adopt CosFace in our \faceloss.}

%We also compare the CosFace loss and ArcFace loss and find that the performance of \frameworkname{} with CosFace loss was higher. The margin parameter $\mu$ plays a key role in the margin-based loss, so we conducted an experiment to examine the effect of $\mu$ on the \faceloss{} we used. We evaluated the recall@20 performance on the \closesetname{} set of MSVD by varying $\mu$ from $0.01$ to $0.2$, as shown in Fig.~\ref{fig:margin}. The results indicate that the CosFace-based loss achieved better performance with higher stability than the ArcFace-based loss. As $\mu$ increased from $0.1$ to $0.2$, the variation of the ArcFace-based loss performance was approximately 0.1, while the variation of the CosFace-based loss performance was less than 0.01. These findings demonstrate the superiority of the CosFace-based loss.

\fuen{Furthermore, as shown in Table~\ref{tab:mbk}, we compare the performance of two different backbones adopted in our music encoder. 1) Vggish: the backbone adopted by CMVAE. 2) ResNet-18, a more advanced residual network~\cite{he2016deep}. The performance of ResNet-18 outperforms Vggish at all metrics on both \closesetname~and \opensetname~despite the fact that Vggish is pretrained on Audio Set~\cite{gemmeke2017audio}. However, the results of adopting Vggish in our \frameworkname~framework still outperform other SOTA approaches on \closesetname~set in Table~\ref{tab:method}.} 

%However, the results of adopting Vggish in our \frameworkname~framework still outperform other SOTA approaches, as shown in Table~\ref{tab:method}.

%outperformed Vggish at all metrics on both \closesetname{} and \opensetname{} as shown in Table~\ref{tab:mbk}.
%Note that Vggish is pre-trained on the audio dataset. Nevertheless, using Vggish as the music encoder in the \frameworkname{} architecture still outperforms other state-of-the-art methods on \closesetname{} 
%according to the performance in Table~\ref{tab:method}.

\paragraph{Comparison on Youtube-8M~\cite{abu2016youtube}.}\label{eva_yt8m}
% \ys{
\fuen{We also evaluate our method on the Youtube-8M, a cross-modal video and music retrieval benchmark dataset, and compare it with SOTA approaches. Our evaluation is performed on the 1024 video-music pairs given by the repository of CEVAR. Since the features provided by Youtube-8M are one-dimensional, we use the original CEVAR backbone instead of R(2+1)D and 2D ResNet-18. The backbone consists of a set of fully connected layers that transform the original features into embeddings, with each hidden layer using ReLU as the activation function. Our results, presented in Table~\ref{tab:table_yt8m}, demonstrate the effectiveness of \frameworkname{} for cross-modal video and music matching on both \closesetname{} and \opensetname{}. Compared to CEVAR, our approach improves Recall@10 by a factor of $2$ on \opensetname{}. These results convincingly demonstrate better applicability of our proposed method than other methods.}
% }

\begin{table}[h]  
\begin{center}  
\resizebox{\columnwidth}{!}{\begin{tabular}{l|cccc}  
% \resizebox{\textwidth}{!}{\begin{tabular}{l|cccc}  
  
\toprule % Toprule applied here  
  
& \multicolumn{4}{c}{\opensettable}\\

& Recall@1 & Recall@5 &Recall@10 & Recall@20 \\  
  
\midrule % Midrule applied here  
   
 \textbf{\frameworkname} & \textbf{0.0645} & \textbf{0.0879} & \textbf{0.1250} & \textbf{0.1807} \\  
 \midrule % Midrule applied here 
 Random & 0.0010 & 0.0049 & 0.0098 & 0.0195\\
 % \midrule % Midrule applied here 
 CEVAR~\cite{suris2018cross} & 0.0105 & 0.0356 & 0.0621 & 0.1041\\ 
 CMVAE~\cite{yi2021cross} & 0.0039 & 0.0147 & 0.0273 & 0.0469\\
  
\bottomrule % Bottomrule applied here  
  
\end{tabular}}%
\caption{Comparison between the proposed \frameworkname{} with state-of-the-art method on Youtube-8M~\cite{abu2016youtube}. Because the video features and music features in Youtube-8M~\cite{abu2016youtube} are one-dimensional, the video encoder and music encoder used in this experiment are fully connected layers. }  
\label{tab:table_yt8m}  
\end{center}  
\end{table}  
\paragraph{Qualitative Comparison.}
\label{qua}
% We visualize the top-5 similarity score of matching music for the video in the \opensetname{} set to further compare \frameworkname{} and CEVAR, as shown in Fig.~\ref{fig:quialitative}. 

\begin{figure*}[t]
    \begin{center}
        \begin{tabular}{c}
             \includegraphics[width=1.0\linewidth]{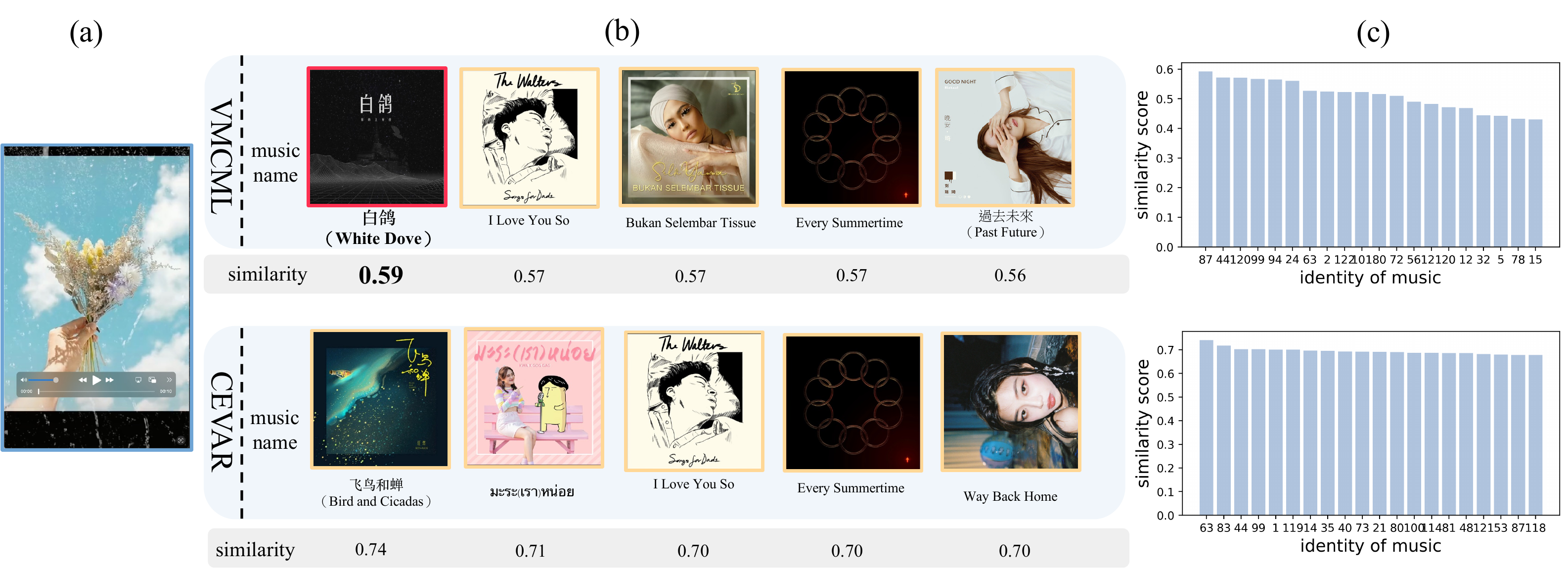}  \\
             \includegraphics[width=1.0\linewidth]{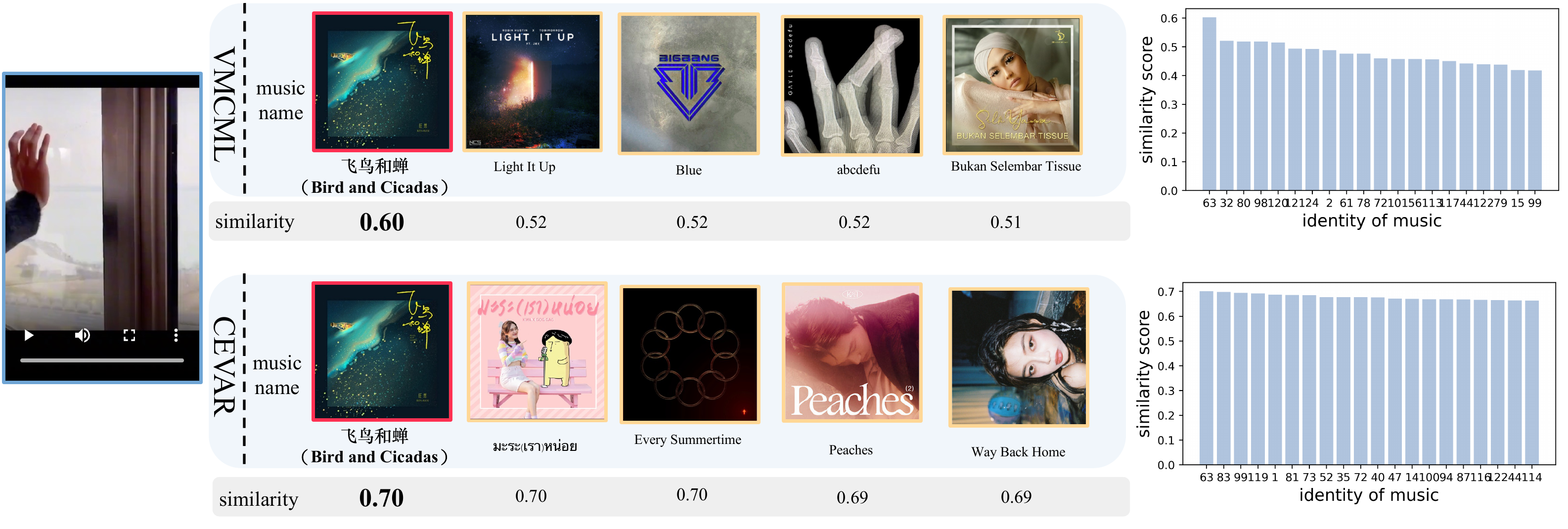} 
        \end{tabular}
    \end{center}
    \caption{
        \textbf{Qualitative of \frameworkname{} for matching video and music}
            is visualized by two examples in \opensetname{} set, and the comparison method CEVAR is shown. (a) are videos (in the blue box). (b) are the top five matching music (in the yellow box) ranked by their similarity score the video, with the ground truth is in the red box. (c) are the top-20 similarity score. 
    }
    \label{fig:quialitative}
\end{figure*}

\fuen{To further address the performance difference between our \frameworkname~and CEVAR, we visualize the predicted top-5 matched music in \opensetname{} for video in Fig.~\ref{fig:quialitative} on \ourdataset~dataset.
The matched music is sorted based on similarity scores in descending order, with the ground truth one highlighted by red boxes. From the first example, we can observe that the \frameworkname{} method predicts the ground truth music with the highest similarity score, while CEVAR cannot even find the ground truth one in the top-5 matched list. 
Also, by visualizing the similarity of the prediction shown in Fig.~\ref{fig:quialitative} (c), we found that the similarity scores of CEVAR for each music only vary in a small range, while \frameworkname~produces higher variances between different music.}

%we find that the CEVAR doesn't spread out the similarity scores for each music.
%In contrast, \frameworkname{} learns to discriminate the features better so that the similarity scores become distinct and meaningful.

% From part (c) Fig. of the second example, we find that the CEVAR doesn't spread out the similarity scores for each music, which means \frameworkname{} learns to discriminate the features better.

% \begin{figure*}[t]
%     % (a) \hspace{85mm}  (b) \hspace{75mm} (c)
%     \small
%     \centering
%     \footnotesize
%     \begin{tabular}{c c c}
%         \textbf{(a)} & \textbf{(b)} & \textbf{(c)}\\
%         \multirow{2}{*}{\includegraphics[height=0.16\linewidth]{src/qua_a_1.pdf}} & \includegraphics[ height=0.16\linewidth]{src/qua_b_1.pdf} & \includegraphics[height=0.16\linewidth]{src/qua_c_1.pdf}\\
%          ~& \includegraphics[height=0.16\linewidth]{src/qua_b_2.pdf} & \includegraphics[height=0.16\linewidth]{src/qua_c_2.pdf}\\
%         \multirow{2}{*}{\includegraphics[height=0.16\linewidth]{src/qua_a_2.pdf}} & \includegraphics[height=0.16\linewidth]{src/qua_b_3.pdf} & \includegraphics[height=0.16\linewidth]{src/qua_c_3.pdf}\\
%          ~& \includegraphics[height=0.16\linewidth]{src/qua_b_4.pdf} & \includegraphics[height=0.16\linewidth]{src/qua_c_4.pdf}\\
%     \end{tabular}
    % \caption{
    %     \textbf{Qualitative of \frameworkname{} for matching video and music}
    %         is visualized by some examples in \opensetname{} set. (a) are videos (in the blue box). (b) are the top five matching music (in the yellow box) ranked by their similarity score the video, with the ground truth is in the red box. (c) are the top-20 similarity score. 
    % }
%     \label{fig:quialitative}
% \end{figure*}

\subsection{Ablation Study}
\label{ablation}
\paragraph{Low-Level Feature.}
\fuen{We conduct an ablation study for the effectiveness of low-level music features extracted from openSMILE feature extraction toolkit, including CHROMA, loudness, and pitch. We compared the video-music matching performance with and without low-level features on both \closesetname{} and \opensetname{}. The results are presented in Table.~\ref{tab:table_llf}. The results suggest that low-level features are crucial for video-music matching tasks. Particularly on \opensetname{}, the performance of Recall@5 is improved by 1.8 times due to the fact that the pitch of the music provides information about the tempo and speed, which are directly associated with the alignment between actions in videos with the drum beats in the music. Additionally, the changes in music volume with time are also an important cue for matching video and music, which further improves the matching performance.}
\begin{table}[h]
    \begin{subtable}[h]{0.45\textwidth}
        % \caption{\closesettable}  
        % \label{tab:table_llf_w}  
        \resizebox{\columnwidth}{!}{\begin{tabular}{l|cccc}  
          
        \toprule % Toprule applied here  
        & \multicolumn{4}{c}{\closesettable}\\
        & Recall@1 & Recall@5 & Recall@10 & Recall@20\\  
          
        \midrule % Midrule applied here  
           
        w/ llf& \textbf{0.2056} & \textbf{0.3562} & \textbf{0.4409} & \textbf{0.5450}\\  
        
        w/o llf& 0.1776 & 0.3109 & 0.3939 & 0.5042\\   
          
        \bottomrule % Bottomrule applied here  
          
        \end{tabular}}% 
    \end{subtable}
    \hfill
    \vspace{10pt}
    \begin{subtable}[h]{0.45\textwidth}
        % \caption{\opensettable}  
        % \label{tab:table_llf_s}  
        \resizebox{\columnwidth}{!}{\begin{tabular}{l|cccc}  
          
        \toprule % Toprule applied here  
        & \multicolumn{4}{c}{\opensettable}\\
        & Recall@1 & Recall@5 & Recall@10 & Recall@20\\  
          
        \midrule % Midrule applied here  
           
        w/ llf& \textbf{0.0170} & \textbf{0.0678} & \textbf{0.1329} & \textbf{0.2526}\\  
        
        w/o llf& 0.0147 & 0.0381 & 0.1079 & 0.2087\\  
          
        \bottomrule % Bottomrule applied here  
          
        \end{tabular}}%
     \end{subtable}
     \caption{Compare \frameworkname{} performance on both seen and unseen music sets when training with and without music \textbf{l}ow-\textbf{l}evel \textbf{f}eatures (\textbf{llf}).}
     \label{tab:table_llf}
\end{table}
% \subsection{Ablation - The effectiveness of similarity loss}\label{esl}

\paragraph{Similarity Loss.}
% We investigate the impact of incorporating the similarity loss, which measures the distance between the output of the video encoder and that of the music encoder. In particular, we consider an anchor video, along with a positive pair of music and a negative pair of music. In addition to learning to distinguish between prototype samples, it is also crucial to learn the differences between sample-to-sample features.
% Furthermore, using multiple layers to learn these differences can enhance performance. We present the results of our experiments in Table~\ref{tab:table_emb}. Our results show that incorporating the similarity loss improves performance on both \closesetname{} and \opensetname{}, indicating the effectiveness of this technique.
\fuen{The effect of \simloss~is shown in Table~\ref{tab:table_emb}. The similarity loss aims at discriminating the features between a positive and a negative pairs of video and music inputs, and consider modality-to-modality distances for further improving matching performance. Hence, our results show that incorporating the similarity loss improves performance on both \closesetname{} and \opensetname{}, indicating that video-music matching task can benefit from this technique.}

%We investigate the effect of incorporating the similarity loss, which is measured in the output of the video encoder and music encoder. The similarity loss learns to discriminate the features from a positive pair and a negative pair of an anchor, which is sample-to-sample and is different from sample-to-prototype such as the face-loss. Furthermore, adding more connections to learn these discriminations can enhance performance. We present the results of our experiments in Table~\ref{tab:table_emb}. Our results show that incorporating the similarity loss improves performance on both \closesetname{} and \opensetname{}, indicating the effectiveness of this technique.
\begin{table}[h]
    \begin{subtable}[h]{0.45\textwidth}
        % \caption{\closesettable}  
        % \label{tab:table_emb_w}  
        \resizebox{\columnwidth}{!}{
        \begin{tabular}{l|cccc}  
          
        \toprule % Toprule applied here  
        & \multicolumn{4}{c}{\closesettable}\\
        & Recall@1 & Recall@5 & Recall@10 & Recall@20\\  
          
        \midrule % Midrule applied here  
           
        w/ $L_{SL}$& \textbf{0.2056} & \textbf{0.3562} & \textbf{0.4409} & \textbf{0.5450}\\ 
        w/o $L_{SL}$& 0.1958 & 0.3206 & 0.3935 & 0.4840\\  
          
        \bottomrule % Bottomrule applied here  
          
        \end{tabular}
        }
    \end{subtable}
    \hfill
    \vspace{10pt}
    \begin{subtable}[h]{0.45\textwidth}
        % \caption{\opensettable}  
        % \label{tab:table_emb_s}  
        \resizebox{\columnwidth}{!}{\begin{tabular}{l|cccc}  
          
        \toprule % Toprule applied here  
        & \multicolumn{4}{c}{\opensettable}\\
        & Recall@1 & Recall@5 & Recall@10 & Recall@20\\  
          
        \midrule % Midrule applied here  
           
        w/ $L_{SL}$& \textbf{0.0170} & \textbf{0.0678} & \textbf{0.1329} & \textbf{0.2526}\\  
        
        w/o $L_{SL}$& 0.0123 & 0.0596 & 0.1166 & 0.2186\\  
          
        \bottomrule % Bottomrule applied here  
          
        \end{tabular}}%
     \end{subtable}
     \caption{Comparison of the performance when training with and without similarity loss on \frameworkname{}.}
     \label{tab:table_emb}
\end{table}
\section{Conclusion}

    In this work, we develop a cross-modal framework \frameworkname{}, which addresses the challenges in music recommendation for new users or new music give short-form videos. \frameworkname{} constructs a shared embedding space between video and music representations effectively by adopting CosFace loss based on margin-based cosine similarity loss. Also, we collect a large-scale dataset (\textbf{\ourdataset{}}) which contains 390 individual music clips and the corresponding matched 150,000 videos, and we will release the dataset after the acceptance. We demonstrate that our approach achieves state-of-the-art performance for matching video and music on \textbf{Youtube-8M} and our \textbf{\ourdataset{}} datasets.

    % We developed a cross-modal framework \frameworkname{} on content-based system for matching video and background music, which addresses the challenges in music recommendation for new users or new music give short-form videos. The \frameworkname{} finds a shared embedding space between video and audio representations effectively by leveraging cosface loss based on margin-based cosine similarity loss.
    % Furthermore, we establish a large-scale dataset called \textbf{\ourdataset{}}. It provides 390 individual music clips and the corresponding matched 150,000 videos. The experiments on \textbf{Youtube-8M} and our \textbf{\ourdataset{}} datasets demonstrate the effectiveness of our proposed framework and achieve state-of-the-art video and music matching performance. %In future work, we would like to ~~~.

{\small
\bibliographystyle{ieee_fullname}
%\bibliography{mybib/conference,mybib/others}

}

\end{document}